# MANet: Multimodal Attention Network based Point-View fusion for 3D Shape Recognition

Yaxin Zhao, Jichao Jiao*, Tangkun Zhang, Xinping Chen, Chenxu Wang, Wei Cui
School of Electronic Engineering
Beijing University of Posts and Telecommunications
Beijing, China
Email: {jiaojichao, zhaoyaxin}@bupt.edu.cn

*Abstract*—3D shape recognition has attracted more and more attention as a task of 3D vision research. The proliferation of 3D data encourages various deep learning methods based on 3D data. Now there have been many deep learning models based on point-cloud data or multi-view data alone. However, in the era of big data, integrating data of two different modals to obtain a unified 3D shape descriptor is bound to improve the recognition accuracy. Therefore, this paper proposes a fusion network based on multimodal attention mechanism for 3D shape recognition. Considering the limitations of multi-view data, we introduce a soft attention scheme, which can use the global point-cloud features to filter the multi-view features, and then realize the effective fusion of the two features. More specifically, we obtain the enhanced multi-view features by mining the contribution of each multi-view image to the overall shape recognition, and then fuse the point-cloud features and the enhanced multi-view features to obtain a more discriminative 3D shape descriptor. We have performed relevant experiments on the ModelNet40 dataset, and experimental results verify the effectiveness of our method.

## I. INTRODUCTION

With the gradual popularization of 3D sensing hardware devices, many 3D applications are on the rise, and 3D visual data such as volume data, multi-view data, and point-cloud data are also attracting more attention. How to understand and deal with 3D data become a hot issue at the moment. 3D shape recognition has become a task in the field of computer vision. It has broad application prospects in many fields such as AR / VR, robots, human-computer interaction, remote sensing mapping and plays an important role in unmanned driving, 3D medical imaging and indoor mapping. In the past, CNN has developed rapidly and achieved state-of-the-art results in the field of 2D image processing. However, due to the sparseness of 3D point-cloud data, it is not very effective. For example, existing methods for processing volume data [1] [2] [3] often bring huge computational cost and poor results. Multi-view data and point-cloud data are popular in 3D data. The former is a 2D image collection captured from different simulation perspectives [4]. In this method, 3D geometric structure can be described in a simple way, but due to the lack of depth information, multi-view images cannot interpret 3D spatial geometric features as well as point-cloud data. Although 2D CNN has been proven to deal with multi-view data effectively, in fact, the multi-view representation requires multiple images to jointly build a complete 3D model. In addition, due to the influence of camera angle, multi-view data will inevitably lose part of information, leading to a lack of global information awareness, or will emerge an overlapping area between adjacent images, which will bring information redundancy and increase unnecessary calculation cost. Point-cloud data are very suitable for 3D scene understanding. After the Lidar scanning, point-cloud can be obtained directly, and the original data can be used for end-to-end deep learning and fully exploring the inherent relationships contained in the data. The biggest advantage of point-cloud is that it can well maintain geometric information in 3D space, but because the data belong to the non-Euclidean domain, which are disordered and sparse, there is no way to apply convolution operations like regular 2D images. At present, there are many mainstream frameworks such as Pointnet [5], Pointnet ++ [6], DGCNN [7] can directly extract features which are more consistent with the constraints of 3D space from point-cloud data, and these methods have achieved competitive results. However, these frameworks do not further exploit fine-grained local features of point-cloud. In recent years, graph-based attention mechanism has been widely used in the field of 3D point-cloud sensing field. [8] proposed GAPNet (Graph Attention based Point Neural Network) to learn local geometric representations that are helpful towards contextual learning by embedding graph attention mechanism within stacked MLP (Multi-Layer-Perceptron) layers.

In this paper, we propose a multimodal attention network called MANet (Multimodal Attention Network) based on point-view fusion. The feature extraction part of our network is divided into point-cloud branch and multi-view branch. First, in the point-cloud branch, graph attention mechanism is used to obtain the high-level global point-cloud features. It is worth noting that, unlike the original GAPNet classification branch, we make some important improvements in the feature extraction process. Inspired by CBAM [9] in 2D convolutions, we extend it to the spatial transform network and feature extraction backbone network based on GAPNet. We want to further explore the relationship between nodes and neighborhood points and name it VNAM (Vertex and Neighborhood Relation Attention Module). Experiments have proved that VNAM can adapt to the architecture of GAPNet and improve the classification accuracy. Second, we believe that there may be overlapping areas between adjacent images of multi-view data, which lead to information redundancy and increase unnecessary calculation cost. In order to use the global point-cloud features to mine the contribution of each multi-view image to the whole shape recognition, we add soft attention mechanism to the point-view fusion module. A soft attention fusion module is designed to generate corresponding attention weights for the different images in multi-view data through the global point-cloud features. Such view attention weights can be used to obtain enhanced multi-view features as we expected, so as to retain the distinguish features and weaken the useless features. Here the residual connection [10] is adopted to obtain the final enhanced multi-view features. Finally, we fuse the global point-cloud features and the enhanced multi-view features to obtain final fusion features, and then the network will connect to the classifier to acquire the final category score. We evaluated the performance of the proposed framework on the ModelNet40 dataset, and the experimental results show that our framework is superior to existing methods of point-cloud-based or multi-view-based as shown in Table I. The evaluation results show that the fusion framework based on multiple attention mechanisms can obtain powerful 3D shape descriptors.

The main contributions of this paper are summarized as follows:

- We propose the first neural network framework named MANet which uses multi-modal attention mechanism to well fuse multi-source data. The feature extraction part of our framework is compatible. First, MANet is based on graph attention mechanism to extract the global point-cloud features, and MANet introduces the soft attention mechanism to use the global point-cloud features to generate corresponding attention weights for different view images. Multi-information fusion is used for 3D shape recognition to achieve high-precision 3D object recognition.

- In the point-cloud branch, a generalized processing method is introduced to explore the attention relationship between nodes and neighborhood points in the 3D point-cloud. We call this processing method VNAM and extend it to the spatial transform network and the backbone network of point-cloud branch. VNAM is used to construct graph features based on graph attention mechanism and improves network performance.

- A soft attention fusion scheme based on point-view data is proposed and the point-cloud global features are used to mine the contribution of each multi-view image to the whole shape recognition. The fusion block will generate a soft attention mask corresponding to each image in the multi-view data and output the final enhanced multi-view features by using residual connection. This method maximizes the use of multi-view features and also combines the two different features of point-cloud and multi-view. Experiments show that this solution is simple and effective.

## II. RELATED WORK

### A. Models Based on Multi-view Data

In the multi-view-based method, 3D geometry is represented by a set of 2D images. Therefore, this method essentially uses the existing deep learning models to perform feature learning on the image collections. At present, there are many frameworks with superior performance can be used in the 2D domain, and different from 3D field, these frameworks can be pre-trained on a massive datasets. For example, MVCNN [4] proposed by Su et al. is based on AlexNet to learn image features from 12 different viewpoints. The features extracted are followed by the view-pooling and then sent to the next CNN to obtain the final shape descriptor. Kalogerakis et al. [11] learn 3D shape features of the shadow map and depth map by FCN. Shi [12] and Sinha et al. [13] use CNN to learn features from 2D images. The former inputs multiple panoramic images of the 3D shape which obtained by cylindrical projection along the major axis direction, while the latter first maps 3D shape to spherical surface, and then projects it to an octahedron to obtain a 2D plane expansion diagram. However, during the transformation process, these methods will inevitably change the original features and lose a considerable amount of geometric structure information.

### B. Models Based on Point-cloud Data

In the point-cloud-based method, 3D shape is represented by a series of disordered point sets distributed in 3D geometric space. In addition to 3D spatial coordinates, these points can also be given other attributes, such as RGB color and reflection intensity. PointNet [5] proposed by Qi et al. is the first method to deal with the irregular 3D data. Max-Pooling is the earliest method to solve the disorder of point-cloud and this method is also widely used in subsequent research. Pointnet++ [6] can extract local features at different scales to obtain deep features through a multi-layer network structure thus solving the problem that Pointnet cannot fully extract local features. Pointsift [14] is designed on the basis of Pointnet ++. It extracts features from 8 important directions in the point-cloud through direction coding units and increases accuracy. However, it increases the calculation cost, causing issues of efficiency and speed.

Recently, graph neural network [15] has returned to people's vision. CNN has a good ability to process the regular 2D data, but it cannot effectively migrate to 3D point-cloud. However, Graph neural network is an ideal choice to process unstructured data. DGCNN [7] proposes a new neural network module named EdgeConv which uses graph-based deep learning methods to extract spatial structured features in point-cloud. It captures the geometric relationship between points by constructing point and its nearest neighbor points into a graph, but it ignores vector direction between adjacent points, and eventually loses a part of local geometric information. GAPNet [8] introduces the graph attention mechanism and introduces self-attention and neighborhood attention mechanisms. It extracts the fine-grained local features of point-cloud in a manner of attention. In order to facilitate contextual understanding, GAPNet uses a multi-head mechanism to capture the attention features of context and shows state-of-the-art performance on the ModleNet40 dataset.

## C. Fusion Models Based on Multimodal Data

Whether multi-view data or point-cloud data, a single type of data may have limitations. Therefore, multimodal data can combine to achieve complementary advantages. For multi-view images, 3D point-cloud features can be used to make consistent predictions on multiple viewpoints and for 3D point-cloud, the high resolution of 2D images can be used to achieve more accurate 3D perception. PVNet [16] is the first network to effectively fuse multi-view images and point-cloud data. The network take multi-view data and point-cloud data as input and use multi-view global features to explore the relationship between local features of point-cloud data. PVRNet [17] aims to explore the relationship between point-cloud data and multi-view data. The author believes that in the process of the fusion of two features, the multi-view images with strong discrimination will obtain more powerful shape descriptors and are more conducive to the 3D shape representation.

We believe that in the multi-view images, although each view image contains different information, there may have redundant information between the adjacent images. The point-cloud as a kind of original data can better preserve the 3D spatial information, so we use the global point-cloud features to guide the network for data attention fusion perception. In this way, the data of the two modalities can be more efficiently fused. More Specifically, our soft attention fusion block will generate different attention weight coefficients for different view images to achieve the effect of filtering multi-view features and fusing two types of data. Based on this view, our network is very different from the work in the above paper.

## III. NETWORK ARCHITECTURE

In this section, we will introduce our proposed network MANet in detail. As shown in Fig. 1, this is our network flowchart. We will first extract features from the two types of data separately, and then perform fusion processing.

As shown in Fig. 2, our network is divided into two branches: point-cloud branch and multi-view branch. The two branches are connected by the soft attention fusion block we proposed. In the point cloud branch, in order to capture the global point-cloud features, we improve the feature extraction model based on GAPNet, more specifically, we propose VNAM to deeply explore the relationship between nodes and neighborhood points in the point-cloud. The test shows that VNAM adapts to the GAPNet classification architecture and improves the performance of the whole network. In the multi-view branch, we refer to the work of MVCNN [4] that feeds 12 multi-view images into a series of convolutional neural networks that share the same convolution kernel. In our experiments, the CNN we use are the first 5 convolutional layers of AlexNet [18]. The difference is that we design a soft attention fusion block instead of directly performing view-pooling as MVCNN. At the same time the global point-cloud features are also fed into the fusion block to explore the contribution of each multi-view image to the whole shape recognition, we believe that this processing method can make greater use of features from multi-view images and achieve a preliminary fusion of the two types of data. The soft attention fusion block uses residual connection and finally outputs the enhanced multi-view features. After that, the multi-view features are connected with the global point-cloud features to generate an overall 3D shape descriptor and the network is then connected to the classifier to obtain the final shape category score.

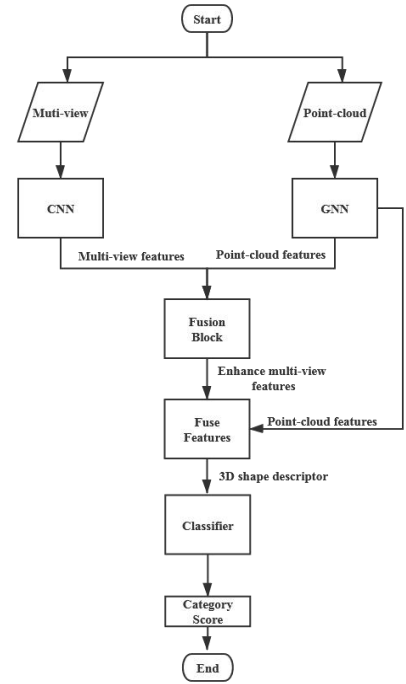

Fig. 1. Network flowchart of the MANet.

## A. Point-cloud Branch and Multi-view Branch

Suppose $X = \{x_i \in R^F, i = 1, 2, \ldots, N\}$ is the input of the point-cloud branch, which is a set of unordered original points. For the sake of simplicity, we set $F = 3$ of the $N$ points, i.e., the 3D coordinates of each point. After comparative consideration, we use the improved GAPNet feature extraction architecture to extract the global point-cloud features. The improved GAPNet uses graph attention mechanism to calculate the self-attention coefficients [26] and the local-attention coefficients, and uses the multi-head mechanism to learn features of each point. The core part of the point-cloud feature extraction architecture is GAPLayer with a multi-head mechanism. As shown in Fig. 2, the input point-cloud data first pass through a spatial transform network with VNAM. The purpose of using attention-aware spatial transform network like this is to maintain the geometric rotation invariance of input data. Feature extraction network uses a 4-head GAPLayer to capture multi-attention features and multi-graph features. We apply VNAM to the part of constructing graph features in spatial transform network and backbone network. More specifically, we add VNAM to each GAPLayer to further improve the performance of the whole network. For the convenience of reader, we show the single-head GAPLayer with VNAM in Fig. 3: a KNN search is used to construct a directed k-nearest neighbor graph and eventually as in Fig.2, the attention feature and graph feature will be fed into the shared MLP and attention pooling layer respectively.

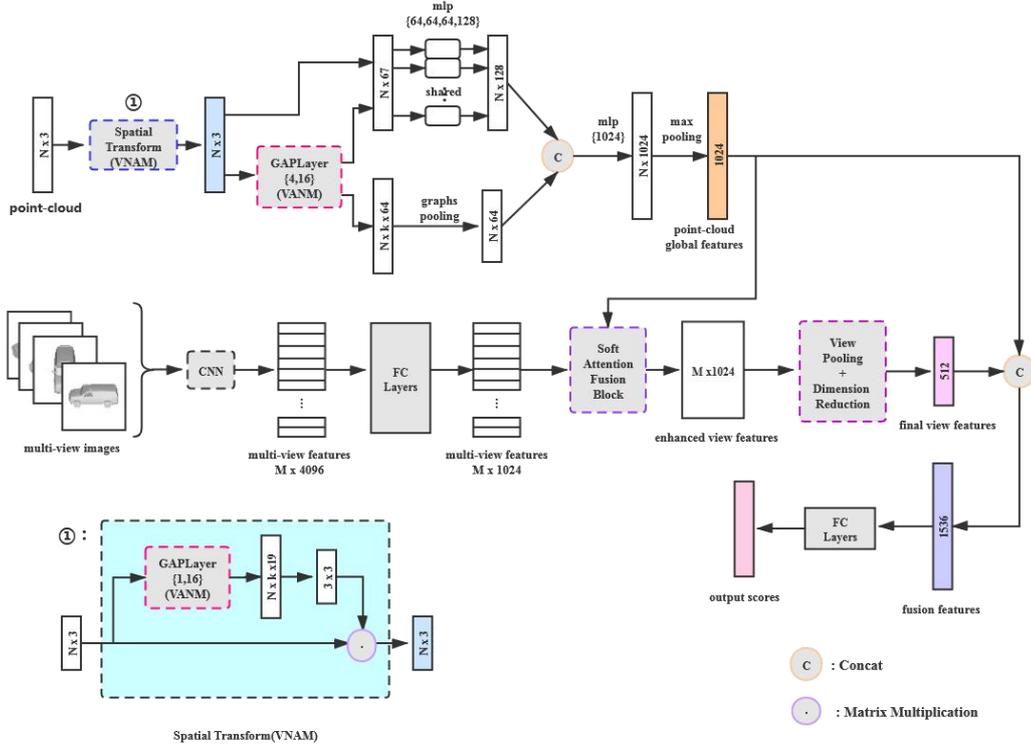

Fig. 2. Architecture of the proposed MANet. The framework is composed of 3 parts: point-cloud branch, multi-view branch and soft attention fusion block. Point-cloud branch: We add the VANM to GAPLayer. In the spatial transform network, GAPLayer{1,16} denotes a GAPLayer with 1 head and 16 channels of encoding feature. we use such GAPLayer to learn a 3×3 transformation matrix. In feature extraction branch, GAPLayer{4,16} denotes a GAPLayer with 4 head and 16 channels of encoding feature. we use such GAPLayer to learn spatial features of the point-cloud.

As shown in Fig. 4, we make some important improvements to point-cloud branch based on the GAPNet architecture. We first introduce VNAM to the 3D point-cloud structure graph. The processing method of VNAM is similar to CBAM [9] in 2D field and VNAM is used to deeply explore the attention relationship between vertex and neighborhood points in the channel and space levels. Each point in the point-cloud is regarded as a vertex represented by $V$ in the directed graph. A directed k-nearest neighbor graph $G = (V, E)$ is constructed to represent the local structure of the point-cloud, where $V = \{x_1, x_2, \cdots, x_n\}$ are nodes for points, $N_i$ is a neighborhood set of point $x_i$ and $E$ represents the edges connecting neighboring pairs of points. In Fig. 4, we define the edge features as $e_{ij} = (x_i - x_{ij}), i \in V, j \in N_i$, where $x_{ij}$ represents the $j$th adjacent point of the point $x_i$.

In the multi-view branch, referring to the existing classic architecture MVCNN [4], we use AlexNet as the basic network and input 12 multi-view images into a series of CNN sharing the same convolution kernel to obtain the multi-view features. In the experiment, our convolutional network uses the first 5 convolutional layers of AlexNet. Here in Fig.2, unlike MVCNN, we did not directly perform view-pooling. Instead, we design a soft attention fusion block to make full use of the content information in each multi-view image. In the fusion block, we use global point-cloud features to perform feature screening on multi-view data to obtain the multi-view features that are more in line with the whole 3D shape recognition.

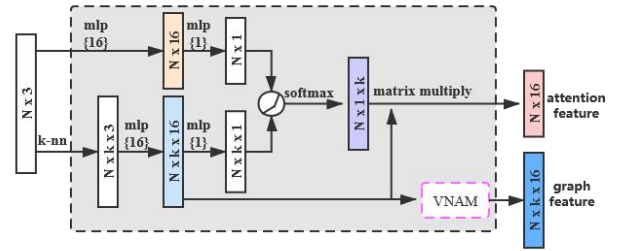

Fig. 3. Single-head GAPLayer with VNAM

### B. Soft Attention Fusion Block

The attention mechanism has recently been brought up again by people and has performed well in many tasks [19] [20] [21] [22]. Among them, the soft attention mechanism is also widely used, which essentially uses a soft information selection mechanism to summarize the input information [23] and unlike the hard attention mechanism [24], the soft attention mechanism can be embedded in deep learning networks for convergence learning. As shown in Fig. 5, soft attention mask is a kind of weight distribution learned by using the similarity between related features. The soft attention fusion block proposed in our paper is a way to fuse different modals data and it can also be understood as the process of feature selection for multi-view features using global point-cloud features. We choose to perform weighted calculations in the feature space and use the global point-cloud feature to generate a soft attention mask for each multi-view image. The learned weight distribution will be element-multiplied with the original multi-view features. The soft attention mask is equivalent to a feature function selector, which enhances useful features in multi-view features and suppresses the features that are not useful for shape recognition. Compared to the method [25] which directly connects features captured from two types of data and then feeds them into the fully connected layer, our processing method can make better use of data features and improve our network's ability to process information.

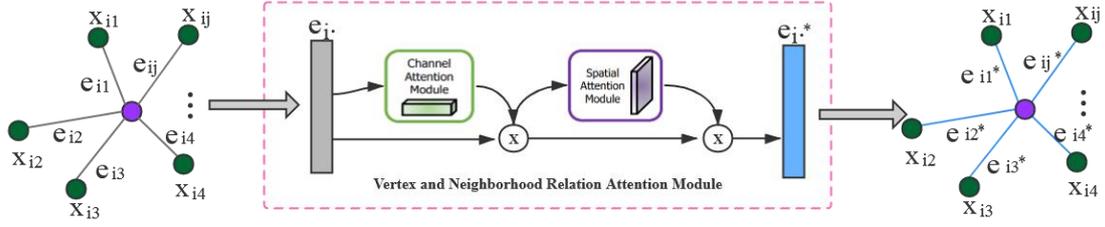

Fig. 4. Schematic diagram of point cloud structure with VNAM module.

As shown in Fig. 5, in the attention fusion block, we follow the residual connection proposed by the Residual Attention Network [10]. First, in order to solve the problem that the two types of features are in different feature spaces, we map the global point-cloud features to the subspace of the multi-view features to obtain the features $P = \{p_1, p_2, \cdots, p_m\}$, and then fuse it with the multi-view features $V = \{v_1, v_2, \cdots, v_m\}$ to obtain the fused features $PV = \{I_1, I_2, \cdots I_m\}$, where m is the number of multi-view images, which we set $m = 12$ in the experiment. Weight coefficients $C(W) = \{w_1, w_2, \cdots, w_m\}$ are generated after normalizing the fused features, i.e.:

$$C(W) = F(MLP(P, V)) \qquad (1)$$

The normalization function $F(\bullet)$ used in this article is the sigmoid function, so the calculation formula of $w_i \in C(W), 1 \leq i \leq m$ is:

$$w_i = \frac{\exp(MLP(p_i, v_i))}{\sum_{k=1}^{m} \exp(MLP(p_k, v_k))}, 1 \leq i, k \leq m \qquad (2)$$

A series of weight distributions $C(W) = \{w_1, w_2, \cdots, w_m\}$ obtained by (2) represent the similarity between each image feature and the global point-cloud features. The attention weight coefficients are element-multiplied with the original multi-view features to obtain the attention weight feature. Finally, referring to the residual connection, the original multi-view features are further added to obtain enhanced features. This method maintains the original information while enhances the multi-view features that are useful for shape recognition. The output of the soft attention fusion block is defined as:

$$O(V^*) = V \bullet (1 + C(W)) \qquad (3)$$

Next, output enhanced features are subjected to the operation of view-pooling and dimension reduction to obtain the final visual shape descriptor which then fuse with the point-cloud global shape descriptor to generate the overall 3D shape descriptor. The second half of the network is connected to a classifier to obtain the final shape category score.

IV. IMPLEMENTATION

Our MANet framework is an end-to-end architecture that introduces the multimodal attention mechanism. Both point-cloud branch and multi-view branch have network compatibility. In our experiments, the low-level multi-view features of the 12 images in the multi-view branch are obtained by the method of MVCNN [4]. The overall architecture of the point-cloud branch is based on the classification framework of GAPNet [8]. We have made some improvements, more specifically, we add VNAM module to the part of computing graph features in spatial transformation network and backbone network. In the soft attention fusion block, we use global point-cloud features to perform feature filtering on multi-view features, and use residual connections to obtain the enhanced multi-view features. After that, we fuse the point-cloud features and enhanced multi-view features again to obtain the final overall 3D shape descriptor. We use a pre-trained multi-view model and point-cloud model to initialize the parameters of our whole network. Inspired by the optimization strategy in PVRNet [17], in the first 10 epochs, we only fine-tune the parameters of our soft attention fusion block and freeze the parameters of the point-cloud and multi-view feature. We will update all parameters after 10 epochs. This optimization strategy is helpful to improve the reliability and stability of the soft attention mask in our fusion block.

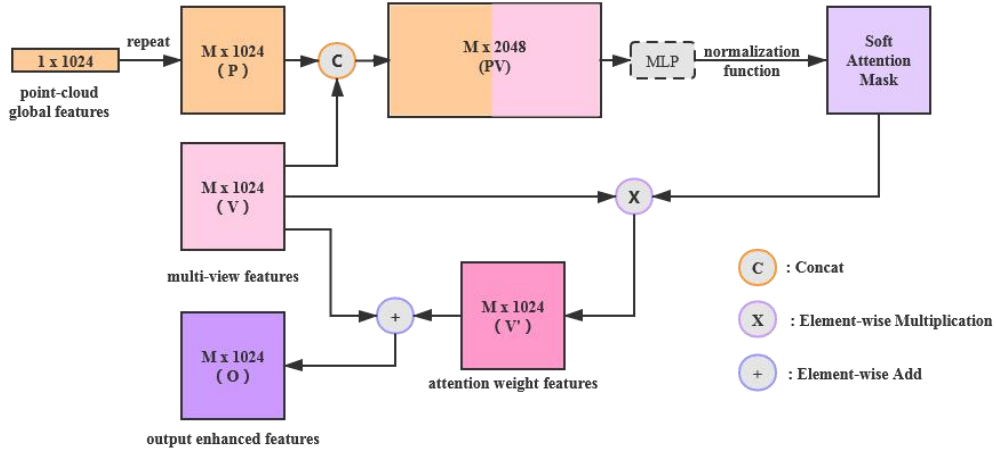

Fig. 5. Soft attention fusion mask. It takes point-cloud global features and multi-view features as input and outputs enhanced multi-view features.

## V. EXPERIMENTS

In this section, we introduce the ModelNet40 dataset, and the performance of our designed network architecture MANet in classification and retrieval. We conduct comparative experiments with existing methods based on multi-view data, point-cloud data and multi-modal fusion data. The part of the visualization results of the soft attention mask in our soft attention fusion block are given and we also perform ablation experiments on the new module VNAM proposed to prove its effectiveness.

### A. Dataset

ModelNet40 [3] is a common subset of the ModelNet dataset, which contains 12,311 CAD models. These CAD models are divided into 40 different categories and we divide 9843 of them into a training set to train our model, and 2468 of them into a test set to verify our model. Our data are divided into point-cloud data and multi-view data. Referring to the work of the Pointnet [5], we also shake the data by randomly rotating and scaling the point-cloud to achieve data enhancement. Multi-view data are borrowed from the related work in MVCNN [4] that observes the CAD model from different simulation perspectives. In the experiment, each CAD model has the point-cloud data and 12 multi-view images which are strictly corresponding.

### B. Experimental Setup

We first use point-cloud data and multi-view data to train two independent feature extraction branches. The point-cloud feature extraction branch is based on the improved GAPNet architecture and the multi-view branch is based on the basic architecture of AlexNet. We train the two branches separately to obtain their own pre-trained models, and then the pre-trained models are used to the overall MANet network training. During our overall network training, we choose the SGD optimizer model with a momentum of 0.9 and set the number of neighbors $k = 20$ in the point-cloud branch, the number of pictures $m = 12$ in the multi-view branch and the batchsize is 20. There are two learning rates in our model, which are used for soft attention fusion block and the overall network. The learning rates have a strategic stepwise adjustment starting from 0.01 and 0.001 respectively. Our model is trained on Pytorch 0.4.1 using 2 NVIDIA TITAN Xp GPUs.

### C. Results

In Table I, we classify the method according to the type of data used, and compare our results with GAPNet [8], DGCNN [7], MVCNN based on AlexNet architecture [4], MVCNN based on GoogLeNet, MVCNN-MultiRes [27] and other recent work. Our model achieves competitive performance on the benchmark of ModelNet40 and is superior to the performance of GAPNet and MVCNN which we refer to. The final results show the effectiveness of our classification model for shape classification. In Fig. 6, we show the visualization results of the 6 CAD models in the soft attention fusion block. Each object corresponds to a point-cloud image and 12 multi-view images. The point-cloud image is a visualization picture generated at a certain angle. The soft attention mask of each multi-view image is displayed below the image.

As shown in Table I, in order to show the effectiveness of our network MANet, we have compared our MANet with the methods based on volume data, multi-view data, point-cloud data, and fusion data. Among them, the volume-based method includes 3D ShapeNets [3], VoxelNet [1], and MVCNN-MultiRes [27]. In the multi-view-based method, we choose 12 multi-view images and MVCNN (GoogLeNet) means that GoogLeNet is employed as base architecture for weight-shared CNN in MVCNN. The point-cloud-based method includes PointNet [5], PoineNet ++ [6], KD-Network [28], SO-Net [29], DGCNN [7] and GAPNet [8].The method based on multimodal data includes FusionNet [30] and PVNet [16]. As shown in Table I, we can see that our architecture MANet classification accuracy is 93.4% and retrieval mAP is 90.1%. In the experiments, we find that our model can achieve 100% recognition accuracy on up to 10 categories and also have 99% accuracy on objects such as bed and monitor. It can be seen

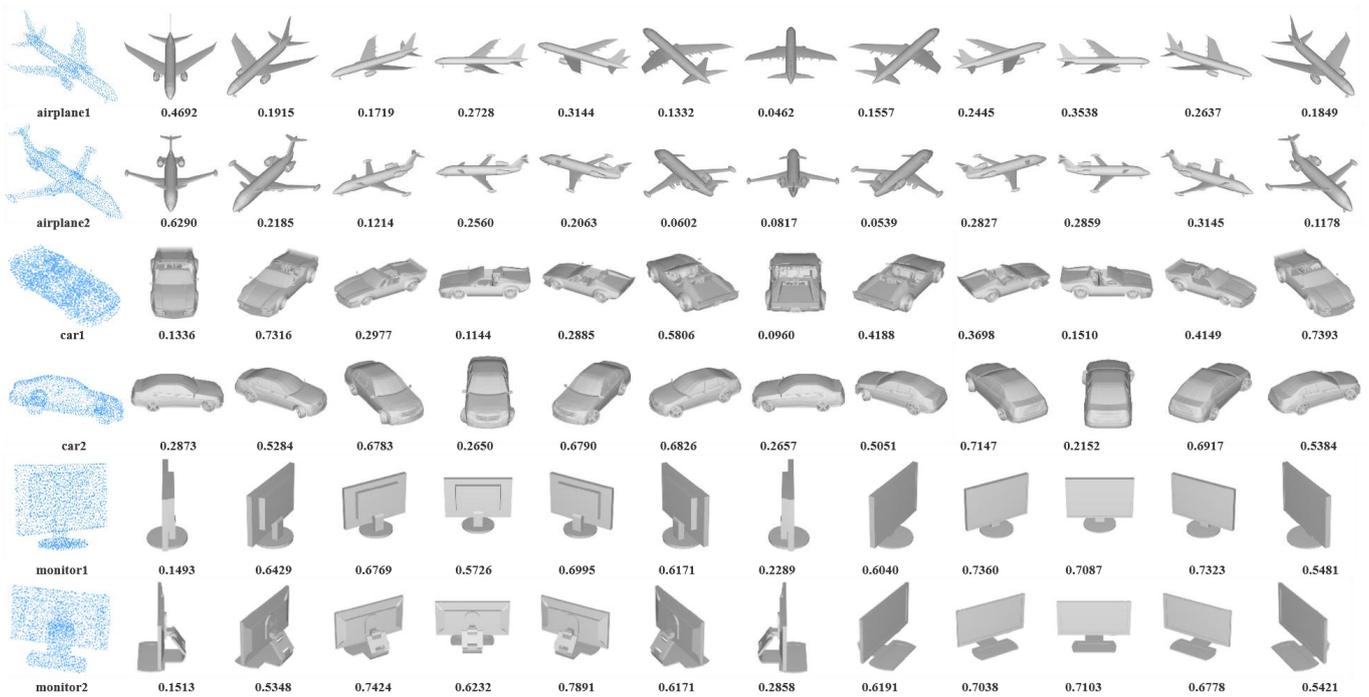

Fig. 6. The visualization results of the soft attention fusion block. The numbers in each line represent the soft attention mask corresponding to the multi-view images.

TABLE I. CLASSIFICATION AND RETRIEVAL RESULTS ON THE MODELNET40 DATASET

| Method | Data | Classification (Overall Accuracy) | Retrieval (mAP) |
|---|---|---|---|
| 3D ShapeNets | Volumetric | 77.3% | 49.2% |
| VoxNet | Volumetric | 83.0% | -[a] |
| MVCNN-MultiRes | Volumetric | 91.4% | - |
| MVCNN(AlexNet) | 12 views | 89.9% | 80.2% |
| MVCNN(GoogLeNet) | 12 views | 92.2% | 83.0% |
| PointNet | Point-cloud | 89.2% | - |
| PointNet++ | Point-cloud | 90.7% | - |
| KD-Network | Point-cloud | 91.8% | - |
| SO-Net | Point-cloud | 90.9% | - |
| DGCNN | Point-cloud | 92.2% | 81.6% |
| GAPNet | Point-cloud | 92.4% | - |
| FusionNet | Volumetric and 20/60 views | 90.8% | - |
| PVNet | Point-cloud and 12 views | 93.2% | 89.5% |
| Ours | Point-cloud and 12 views | 93.4% | 90.1% |

[a.] Symbol '-' Means Results Are Unavailable.

that for objects with obvious features, our model can achieve accurate recognition, but it should be pointed out that the model has a poor recognition rate on objects such as flowerpot and cup. How to make the network distinguish difficult objects accurately will become our future improvement direction.

### D. Vertex and Neighborhood Relation Attention Module

In this section, we will show the improvements we have made in the point-cloud branch. As shown in Table II, our proposed structural VNAM improves the recognition accuracy by 0.24% without affecting the training time and the model size. (It should be noted that the final classification accuracy of GAPNet is 91.21% based on our experimental environment, and the accuracy of the original author is 92.4%, so we have marked it in Table II.)

TABLE II. ABLATION EXPERIMENTS OF VNAM

| Method | Classification (Overall Accuracy) | Model Size |
|---|---|---|
| GAPNet | 91.21% (92.40%) | 22.9 MB |
| GAPNet(CBAM) | 91.45% | 21.9 MB |

### VI. CONCLUSIONS

This paper proposes a fusion network MANet based on multi-modal attention mechanism for 3D shape recognition. MANet is a new neural network that can use the global point-cloud features to filter multi-view features to achieve the fusion of two features. In the multi-view branch, we refer to the related work of MVCNN. In the point-cloud branch, we have improved the GAPNet classification architecture. Here, we propose a VNAM attention module and gain accuracy improvement. In the fusion branch, we design a soft attention fusion block to achieve the processing of multi-view features while fusing the two features. Finally, we got a unified 3D shape descriptor. The classification and retrieval experiments on ModelNet40 show the effectiveness of our framework. In the future, we can further explore the fusion of multiple modal data and the deep application of the attention mechanisms.